# Forecasting Method for Grouped Time

# Series with the Use of k-Means Algorithm

**Nadezhda N. Astakhova**

Ryazan State Radio Engineering University
Gagarin Str., 59/1, Ryazan, Russian Federation, 390005

**Liliya A. Demidova**

Ryazan State Radio Engineering University
Gagarin Str., 59/1, Ryazan, Russian Federation, 390005

Moscow Technological Institute
Leninskiy pr., 38A, Moscow, Russian Federation, 119334

**Evgeny V. Nikulchev**

Moscow Technological Institute
Leninskiy pr., 38A, Moscow, Russian Federation, 119334
&
National Research University Higher School of Economics (HSE)
20, Myasnitskaya str., Moscow, 101000

## Abstract

The paper is focused on the forecasting method for time series groups with the use of algorithms for cluster analysis. K-means algorithm is suggested to be a basic one for clustering. The coordinates of the centers of clusters have been put in correspondence with summarizing time series data – the centroids of the clusters. A description of time series, the centroids of the clusters, is implemented with the use of forecasting models. They are based on strict binary trees and a modified clonal selection algorithm. With the help of such forecasting models, the possibility of forming analytic dependences is shown. It is suggested to use a common forecasting model, which is constructed for time series - the centroid of the cluster, in forecasting the private (individual) time series in the cluster. The promising application of the suggested method for grouped time series forecasting is demonstrated.

**Keywords:** time series, clustering, cluster centroid, k-means algorithm, average forecasting error rate, strictly binary tree, antibodies, antigens, modified clonal selection algorithm

# 1 Introduction

Nowadays development of various socio-economic processes, which are becoming increasingly interconnected, accelerates more actively. In this regard, it is necessary to use approaches, which are already well known, more extensively. It is also important to develop new approaches to solving the problem of forecasting grouped time series (TS), which describe certain process indicators.

The plain fact is that during the development of forecasting TS models it is necessary to consider not only the tendencies of forecasted TS, but also tendencies of other TS which can influence on it. For example, such macroeconomic indicators as gross domestic product and the level of exports described by a TS are in explicit dependence, as well as the average level of income and level of education, health care and average life expectancy.

At the same time it is also clear that many TS have similar laws of changing values of its components and they can be combined in a subgroup according to the criterion of similarity of these laws [12]. In this case, it is advisable to develop a forecasting model that is common to all TS included in the subgroup for the purpose of direct use to forecast all private TS subgroups or further refinement of forecasting models for individual private TS (in the case of lack of precise predicted values while using the general model) [11].

The process of developing the model of forecasting TS is characterized by the highest computational complexity. That means that the use of a common forecasting model for a TS subgroup can be a significant step in the development of approaches to the analysis of correlating TS. It allows to carry out forecasting of private TS in a subgroup with acceptable time-consuming.

It is obvious that the use of cluster analysis techniques, namely clustering algorithms such as $k$-means algorithm [8, 9], the algorithm of fuzzy $c$-means (FCM-algorithm) [2, 3], EM-algorithm [10], will bring together similar TS into subgroups (clusters) and determine TS-centroids of clusters. All of that is necessary for solving the problem of common forecasting models development for TS subgroups. In addition to actual development of forecasting TS-centroid clusters models, it is suggested to use an approach that supposes construction of forecasting models on the base of strict binary trees (SBT) and modified clonal selection algorithm (MCSA) with the formation of analytical dependencies. They describe certain values of TS and provide minimum affinity values, in other words average forecasting error rate [6].

The aim of the paper is to develop an approach to forecasting grouped time series with the use of cluster analysis algorithms, which provides a forecast with reasonable time-consuming.

## 2 Theoretical part

Suppose that there is a group of TS $T$ : $t_i$ $(i = \overline{1, m})$. Therewith, let each TS $t_i$ consists of $n$ $(10 \le n \le 30)$ elements $t_i^j$ $(j = \overline{1, n})$.

Significant interest is in the solution of the problem of developing forecasting models for all TS in a group with reasonable time-consuming.

Time series, as well as other objects of data analysis, can be joined into clusters (subgroups) taking into account values of certain features. They are expectation values, dispersion and others. They can be calculated on the base of TS elements values, maximum and minimum values of TS, etc.

The problem of selection of an analytic dependence is usually solved in the process of forecasting models development for the purpose of calculating TS future values. This analytic dependence describes the variation of TS values of the elements in time in a best way. So, it seems appropriate to use similar mathematical laws of changing TS values in time for grouping TS into clusters.

The hypothesis of possible similarities between mathematical laws of changing TS values, and consequently, correspondent kinds of analytical dependences can be explained by the fact that many socio-economic indicators are interrelated, and changing trends of one of them causes a change in the trends of the other.

Unfortunately, there is no available information about what TS form connected subgroups in a group of analyzed TS. This fact leads to a need for additional methods of analysis, such as cluster analysis techniques.

The use of such clustering algorithms as *k*-means algorithm [8, 9], FCM-algorithm [2, 3], EM-algorithm [10], during iterative calculations allows to divide groups of objects into a predetermined number of clusters according to some optimality criterion, in addition to determining the coordinates of clusters centroids.

During handling the problem of TS clustering, centroids coordinates can be used to form integrating TS-centroids, which characterize private TS included in corresponding clusters. It is obvious that a certain forecasting model can be developed for TS centroid. Then it can be used either directly for forecasting private TS assigned to a particular cluster or as a basic model for the purpose of further clarification and application for forecasting private TS.

In order to minimize the time spent on the development of forecasting models, it is suggested to use *k*-means algorithm [9] that implements the partition of objects of group $T$ into sub-groups $T_r$ $(r = \overline{1, c})$ so that

$$\bigcup_{r=1}^{c} T_r = T; \tag{1}$$

$$T_r \bigcap T_h = \varnothing; \quad r = \overline{1, c}; \quad h = \overline{1, c}; \quad r \neq h;$$

$$\varnothing \subset T_r \subset T; \quad r = \overline{1, c}.$$

In the context of solving TS clustering problem it is supposed to understand TS as an object. Let $u_r(t_i)$ be a characteristic function, which can take two values: 0, if the object does not belong to the cluster, and 1 if object belongs to a cluster. Then a clear $c$-partition of the group objects in the cluster can be described by the matrix $U = [u_r(t_i)]$ $(u_r(t_i) \in \{0,1\}; \quad r = \overline{1, c}; \quad i = \overline{1, m})$.

The matrix $U = [u_r(t_i)]$ must meet the following requirements:

$$\sum_{r=1}^{c} u_r(t_i) = 1 \quad (i = \overline{1, m}); \tag{2}$$

$$0 < \sum_{i=1}^{m} u_r(t_i) < m \quad (r = \overline{1, c}). \tag{3}$$

K-means algorithm provides minimizing of the objective function:

$$J(U, V) = \sum_{r=1}^{c} \sum_{i=1}^{m} u_r(t_i) \cdot d^2(v_r, t_i), \tag{4}$$

where $U = [u_r(t_i)]$ is the $c$-partition of the group objects $T$ on the base of the characteristic functions $u_r(t_i)$, that define that the object $t_i$ belongs to a cluster $T_r$; $V = (v_1, ..., v_c)$ — centroids of clusters; $d(v_r, t_i)$ is the distance between the center of the cluster $v_r$ and the object $t_i$; $c$ is the number of clusters $T_r$; $m$ is the number of objects; $r = \overline{1, c}; \quad i = \overline{1, m}$.

The following steps should be done during the implementation of $k$-means algorithm [3].

1. Initialization of the initial $c$-partition $U = [u_r(t_i)]$, that must match the requirements (2) and (3).

2. The calculation of clusters centroids coordinates:

$$v_r^j = \frac{1}{m_r} \cdot \sum_{i=1}^{m} u_r(t_i) \cdot t_i^j,$$

where $m_r$ is the number of objects related to cluster $r$; $j = \overline{1, n}$.

3. Calculation of a new $c$-partition $U = [u_r(t_i)]$, that must match the requirements (2) and (3).

4. Steps 2 and 3 are repeated until the desired accuracy $\varepsilon$: $|J(U, V) - J'(U, V)| \leq \varepsilon$, where $J(U, V)$, $J'(U, V)$ are the values of the objective function (4) in two successive iterations (or until prescribed number of iterations $H$ is done).

If it is necessary to calculate the distance $d(v_r, t_i)$ between the center of the cluster $v_r$ and the object $t_i$, usually Euclidean metric is used. In this case $d(v_r, t_i)$ is calculated as

$$d(v_r, t_i) = \left[ \sum_{j=1}^{n} (v_r^j - t_i^j)^2 \right]^{0.5},$$

where $n$ is the features' number of the object. So in the context of solving the problem of TS clustering on the base of similarity of mathematical laws of TS values changing, it is suggested to do modification to perform this metric so as to take into account various relevance of TS elements (most distant in time from the time of forecasting) and larger relevance of other TS elements (closest in time to the moment of forecasting) [1]

$$d(v_r, t_i) = \left[ \sum_{j=1}^{n} \frac{j}{n} \cdot (v_r^j - t_i^j)^2 \right]^{0.5}. \tag{5}$$

The utility of such a modification can be explained so that with the passage of time the dependencies between socio-economic indicators change. So, during the forming of forecasting models, larger preference should be given to the closest TS elements in time of forecasting. The use of the weighting coefficients $j/n$ suggests the most significant differences between the values of the most relevant TS elements (for example, if $j = n$ then the weight value is 1, and if $j = 1$ then it is $1/n$). The use of formula (5) will not only take into account the relevance of TS elements, but also in a view of high sensitivity to divergent TS trends, it will provide unification of clusters on the base of similarity of the trends.

However, there is a problem, the solution of which is a principal in the case of cluster analysis techniques. This problem is related to the different scale of the analyzed TS which characterize various socio-economic indicators. Such indicators have different unit of measurement, a variety of ranges and corresponding statistical features (statistical expectation and so on).

To solve this problem it is advisable to use the algorithms of normalization, which are widely used in the statistics, mathematical economics and econometrics. Their point is to define a medium level — median, according to which all analyzed TS are aligned.

A certain conditional straight can be used as a median, as well as one of TS of the analyzed TS group or TS-centroid $S$, which $S_j$ elements are defined as:

$$S_j = \frac{1}{m} \cdot \sum_{i=1}^{m} t_i^j,$$

where $t_i^j$ is the $j$-th element of the $i$-th TS; $i = \overline{1, m}$; $j = \overline{1, n}$; $m$ is the number of TS; $n$ is the number of elements of TS-centroid.

Normalization algorithm of the $i$-th TS $(i = \overline{1, m})$, which is based on TS-centroid $S$, can be represented by the following sequence of steps.
*Step 1.*
1.1. The average step $hS$ of changing TS-centroid elements is determined as

$$hS = \left( \max_{j=1,n}(S_j) - \min_{j=1,n}(S_j) \right) \Big/ n,$$

where $n$ is the number of elements of TS-centroid; $S_j$ is the $j$-th element of TS-centroid $S$.

1.2. The average step $ht_i$ of changing of the $i$-th TS elements $(i = \overline{1, m})$ is determined:

$$ht_i = \left( \max_{j=1,n}(t_i^j) - \min_{j=1,n}(t_i^j) \right) \Big/ n,$$

where $n$ is the number of TS elements; $t_i^j$ is the $j$-th element of the $i$-th TS.

*Step 2.*

2.1. The average TS centroid level is determined:

$$\overline{S} = \frac{1}{n} \sum_{j=1}^{n} S_j,$$

where $n$ is the number of elements of TS centroid; $S_j$ is the $j$-th element of TS centroid.

2.2. The average level $\overline{t_i}$ of the $i$-th TS $(i = \overline{1, m})$ is determined as:

$$\overline{t_i} = \frac{1}{n} \sum_{j=1}^{n} t_i^j,$$

where $n$ is the number of TS elements; $t_i^j$ is the $j$-th element of the $i$-th TS.

*Step 3.*

The value $dt_i^j$ is calculated. It represents a ratio of diminution between the average level $\overline{t_i}$ of the $i$-th TS and the $j$-th element $t_i^j$ of the $i$-th TS to the average step $ht_i$:

$$\Delta t_i^j = (\overline{t_i} - t_i^j) \big/ ht_i,$$

where $t_i^j$ is the $j$-th element of the $i$-th TS; $\overline{t_i}$ is the average level of the $i$-th TS; $ht_i$ is the average step of changing of the $i$-th TS elements; $i = \overline{1, m}$; $j = \overline{1, n}$.

*Step 4.*

The $j$-th element $t_i^j$ of the $i$-th TS is transformed into:

$$t_i^j = \overline{S} + \Delta t_i^j \cdot hS.$$

Such converted TS can further be used for clusterization with the help of $k$-means algorithm.

Since clusters centroids express general trends for TS subgroups, which form correspondent clusters, it is appropriate to develop forecasting models for TS-centroid.

Nowadays, there are different approaches to building TS forecasting models. One of the most perspective approach is the implementation of evolutionary algorithms (genetic algorithms [13, 14], clonal selection algorithms [4, 7]) which based on the principles of natural selection. They provide (at acceptable time expenses) construction of TS forecasting models that describe in certain TS values a best way. They can be also characterized by acceptable indicators values of quality models.

In the context of solving the problem of forecasting models development for TS-centroid of clusters, it is appropriate to use the modified clonal selection algorithm. This algorithm simulates the natural laws of the immune system functioning [5, 15] and provides the formation of quite complex analytical functions [6]. The principles of developing forecasting models of $k$-order with the use of MCSA were investigated in [7]. MCSA allows to form an analytical dependence on the base of SBT at an acceptable time expenses, that describes certain TS values and provides a minimum affinity (affinity) — average forecasting error rate (AFER):

$$AFER = \left[ \sum_{j=k+1}^{n} \left| (f^j - d^j)/d^j \right| \right] \cdot \frac{1}{n-k} \cdot 100\%, \qquad (6)$$

where $d^j$ и $f^j$ are respectively the actual (fact) and forecasted values for the $j$-th element of TS (for the $j$-th timing); $n$ is the number of TS elements (number of timing).

In the context of solving the problem of forecasting TS subgroups as $d^j$, it is suggested to use, for example, the $j$-th element $t_i^j$ of the $i$-th TS or the corresponding $j$-th element $v_r^j$ of the $r$-th TS centroid cluster.

Possible options for analytical dependences are presented in the form of antibodies $Ab$, which recognize antigens $Ag$ (certain TS values). An antibody $Ab$ is selected as «the best one». It provides the minimum value of affinity $Aff$ [6]. Coding of an antibody $Ab$ is carried out by recording signs in a line. The signs are selected from three alphabets [6]:

− the alphabet of arithmetic operations (addition, subtraction, multiplication and division) — $Operation = \{'+', '-', '\cdot', '/'\}$;

− the functional alphabet, where letters $S', 'C', 'Q', 'L', 'E'$ define mathematical functions «sine», «cosine», «square root», «natural logarithm», «exhibitor», and the sign '_' means the absence of any mathematical function, − $Functional = \{'S', 'C', 'Q', 'L', 'E', '\_'\}$;

− the alphabet of terminals, where letters $'a', 'b', ..., 'z'$ define the arguments required analytical dependence and the sign '?' defines a constant, $Terminal = \{'a', 'b', ..., 'z', '?'\}$.

The use of these three signs alphabets provides a correct conversion of randomly generated antibodies into the analytical dependence. The structure of such antibodies can be described with the help of SBT [6, 7].

The number of signs in the alphabet of terminals $Terminal$ in the antibody $Ab$ determines maximal possible order $K$ of forecasting models ($K \geq k$, where $k$ is the real model order), i.e. having the value of the element $d^j$ in forecasting TS at the $j$-th moment of time, $K$ values of TS elements can be used as: $d^{j-K}$, …, $d^{j-2}$, $d^{j-1}$.

The use of SBT type, illustrated in Fig. 1, allows to build complex analytical dependence and provides high accuracy of forecasting TS [6].

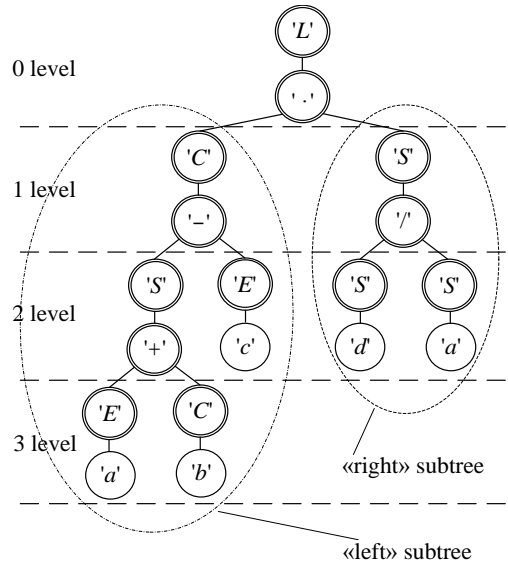

Fig. 1. An example of a strict binary tree, which is used to form antibodies

Such SBT can be generated as a composition result of one «left» subtree of the maximum possible order $K = 3$ and some «right» subtrees of the maximum possible order $K = 2$.

Thus the term «left» subtree («right» subtree) is used for the branch (left or right) of SBT level in which it is necessary to include a new subtree.

In this case it is rational to form antibodies by subdividing SBT into subtrees, then execute the subtree-walk of each vertex forming the ordered symbol lists on its vertices and then combining these lists consecutively [6].

Forming the symbol ordered list on the base of a subtree the consecutive double subtree-walk is carried out: at first moving the subtree bottom-up left to right we walk the vertices containing the alphabetic terminal signs *Terminal* in pairs and correspondingly above placed vertices containing the alphabetic functional symbols *Functional*, and then moving in the same direction it is necessary to go around in pairs the vertices containing the alphabetic arithmetic operation signs *Operation* and correspondingly above placed vertices containing the alphabetic functional signs *Functional*.

The first two signs in such an antibody contain the pair of zero level SBT from the functional alphabet *Functional* and arithmetic operation alphabet *Operation*.

Then there are the lists of the signs corresponding to the «right» maximum possible ordered subtrees $K = 2$ (moving the SBT bottom-up) and finally the symbol list of the «left» maximum possible ordered subtree $K = 3$.

Using such a way of antibody formation we ensure the visualization of the SBT structure representation in the form of the subtrees union, and the antibody is easily interpreted in the analytical dependence.

For example, the antibody formed on the base of SBT as shown in Fig. 1 is coded by the line of signs:

$$L \cdot S \, / \, SeSdC - S + EaCbEa,$$

which can be transformed into an analytical dependence:

$$f(a, b, c, d) = \ln(\cos(\sin(\exp(a) + \cos(b)) - \exp(c)) \cdot \sin(\sin(d) / \sin(a)). \quad (7)$$

In the task of a forecasting model development of $k$-order where $k = 4$ and considering the order of $a, b, c, d$ in the terminal alphabet *Terminal*, the analytical dependence (7) can be written as:

$$f(d^{j-1}, d^{j-2}, d^{j-3}, d^{j-4}) = \ln(\cos(\sin(\exp(d^{j-1}) +$$
$$+ \cos(d^{j-2})) - \exp(d^{j-3})) \cdot \sin(\sin(d^{j-4}) / \sin(d^{j-1})).$$

Interpreting the antibodies into the analytical dependences it is rational to use the recursive procedure of interpretation [6]. MCSA applied to the searching for «the best» antibody defining «the best» analytic dependence includes the preparatory part (realizes the formation of the initial antibody population) and iterative part (presupposes the ascending antibodies ordering of affinity *Aff*; the selection and cloning the part of «the best» antibodies, that are characterized by the least affine value *Aff*; the hypermutation of the antibodies clones; self-destruction of the antibodies clones «similar» to the other clones and antibodies of the current population; calculating the affinity of the antibodies clones and forming the new antibodies population; suppression of the population received; generation of the new antibodies and adding them to the current population until the ingoing size; the conditional test of the MCSA completion.

It is arguable that the application of the general forecasting models developed for TS-centroid of the clusters ensures the uniqueness of the forecasting result for all the particular TS labeled as a cluster since while prognostic value calculation for the particular TS based on the analytic dependences defining prognostic models for TS-centroid of the clusters private item values of TS stand for the values of variables.

On the base of the foregoing approach to a TS group forecasting ensuring the forecasting result within the acceptable expenditures of time can be realized after execution of the following steps.

*Step 1*. Normalization of all TS groups in reference to their average level expressed by the TS-centroid $S$.

*Step 2*. Normalized TS clusterization of the set quantity of clusters $c$.

*Step 3*. Forecasting model creation for the TS-centroid of clusters on the base of SBT and MCSA.

*Step 4*. Private initial TS forecasting with the use of the forecasting models for the TS-centroid of clusters.

It should be noted that the TS forecasting with use of the general forecasting models doesn't conduct to receiving the general forecasting for the TS subgroup (cluster). The forecasting model defines only the mathematical law of elements value change of TS by means of the analytical dependence formed with MCSA applied. The forecasted values for every particular TS will be unique as they will be calculated by means of substitution known private elements values of TS in the general forecasting model.

## 3 Experimental studies

Approbation of the offered approach to forecasting grouped TP was executed with use of TS for 22 macroeconomic indicators of Russian Federation taken from the site World DataBank from 1999 to 2014 (http://databank.worldbank.org/data/views/reports/tableview.aspx?isshared=true#).

All indicators were divided into 4 clusters (subgroups) with use of *k*-means algorithm. Information on the clusters' contents is given in the first column of Table 1.

The general forecasting models for each cluster were defined on the base of antibodies compared to TS-centrodes of clusters:

$$\_-C\cdot\_-\_-C\cdot E/E-Q\,/\,\_bLhCf\,0iSeCaCbQiQf\,;$$
$$L-C\cdot Q\,/\,\_-C\cdot C-E\,/\,Q-EcSeS?QcQ?QfEiCdQc\,;$$
$$L+\_-C-\_-S-Q\cdot C\cdot C\cdot\_iEcEdQhSeEbCbQ?Sg;$$
$$L+Q+Q\cdot E+C\cdot C\,/\,Q+S\cdot Si\,\_bE?Ca\_fC?QcEfCe$$

in the form of the following analytical dependences:

$$f_1(d^{j-1}, d^{j-2}, d^{j-3}, d^{j-4}, d^{j-5}, d^{j-6}) = \cos(\exp(\mathrm{sqrt}(\mathrm{sqrt}(d^{j-3})\,/\,\mathrm{sqrt}(d^{j-1})) -$$
$$-\cos(d^{j-5})) - \exp(\cos(d^{j-6})\,/\,\sin(d^{j-4}))\cdot\cos(d^{j-1}\cdot\cos(d^{j-3}))) - \ln(d^{j-2}) - d^{j-5};$$
$$f_2(d^{j-1}, d^{j-2}, d^{j-3}, d^{j-4}, d^{j-5}) = \ln(\cos(\mathrm{sqrt}(\exp(\mathrm{sqrt}(\mathrm{sqrt}(d^{j-5}) -$$
$$-\cos(d^{j-4}))\,/\,\exp(d^{j-1})) - \cos(\mathrm{sqrt}(d^{j-2}) - \mathrm{sqrt}(25,87)))\cdot\cos(\mathrm{sqrt}(d^{j-5})\cdot$$
$$\cdot\sin(-103,17)))\,/\,\sin(d^{j-3}) - \exp(d^{j-5}));$$
$$f_3(d^{j-1}, d^{j-2}, d^{j-3}, d^{j-4}, d^{j-5}, d^{j-6}) = \ln(\mathrm{sqrt}(\mathrm{sqrt}(\mathrm{sqrt}(\sin(\cos(d^{j-3})\cdot$$
$$\cdot\exp(d^{j-2}) + \mathrm{sqrt}(d^{j-4})) + \cos(\cos(2,14)\,/\,d^{j-2})) + \cos(\cos(d^{j-6})\cdot$$
$$\cdot\exp(224,07)))\cdot\exp(d^{j-5} + \sin(d^{j-1})));$$
$$f_4(d^{j-1}, d^{j-2}, d^{j-3}, d^{j-4}, d^{j-5}) = \ln(\exp(\exp(\exp(\sin(\ln(d^{j-4})\cdot$$
$$\cdot\sin(d^{j-3}))\,/\,\exp(129,43)) - \cos(\ln(d^{j-3}) +$$
$$+\mathrm{sqrt}(d^{j-2})))\cdot\cos(\mathrm{sqrt}(d^{j-5}) + d^{j-1}))\cdot\mathrm{sqrt}(d^{j-4}) - \exp(d^{j-1})).$$

Table 1. Results of macroindicators' forecasting

| No | Indicator name | Measurement unit | *AFER*, % | 2012 | | 2013 | | 2014 | | Error, % |
|----|----------------|------------------|-----------|------|----------|------|----------|------|----------|----------|
| | | | | fact | forecast | fact | forecast | fact | forecast | |
| Cluster 1 | | | | | | | | | | |
| 1 | Power consumption | KOE* | 1,70 | 4740 | 5105,56 | 5022 | 4733,09 | 5114 | 5013,35 | 5,09 |
| 2 | Electricity consumption | kWh/person | 0,74 | 6279 | 6476,5 | 6457 | 6269,51 | 6485 | 6447,64 | 2,21 |
| 3 | Gross national income per capita on the Atlas method | $ | 5,02 | 9642 | 12730,34 | 10406 | 9633,48 | 11740 | 10397,98 | 15,06 |
| 4 | Gross national income per capita at par purchasing power | $ | 3,25 | 19373 | 22700,58 | 20861 | 19363,23 | 22279 | 20852,1 | 9,75 |
| Cluster 2 | | | | | | | | | | |
| 5 | Value added in a services sector | % of GDP** | 0,36 | 59,59 | 58,2 | 59,82 | 59,2 | 59,52 | 59,59 | 1,18 |
| 6 | Export of goods and services | % of GDP** | 0,61 | 29,04 | 30,3 | 29,03 | 29,6 | 29,05 | 29,04 | 2,04 |
| 7 | Import of goods and services | % of GDP** | 0,29 | 21,9 | 21,7 | 21,87 | 22,3 | 21,89 | 21,9 | 0,97 |
| 8 | Gross accumulation of capital | % of GDP** | 0,23 | 24,95 | 25 | 24,91 | 24,5 | 24,84 | 24,95 | 0,77 |
| 9 | Income (except for grants) | % of GDP** | 0,97 | 29,04 | 31,3 | 29,14 | 29,8 | 28,96 | 29,04 | 3,24 |
| 10 | Coefficient of teenage fertility | Births' quantity/ 1000 women | 2,26 | 28,97 | 26,4 | 29 | 25,7 | 28,98 | 28,97 | 7,54 |
| 11 | Value added in the industry | % of GDP** | 2,91 | 34,15 | 37,4 | 35,12 | 36,8 | 36,16 | 34,15 | 6,38 |
| Cluster 3 | | | | | | | | | | |
| 12 | Procedures' start of for business registration | Quantity | 1,48 | 7,95 | 8 | 7 | 8 | 7,82 | 7,95 | 4,92 |
| 13 | Export of high technologies | % | 1,36 | 8,27 | 8 | 8,95 | 8,4 | 8,57 | 8,27 | 4,52 |
| 14 | Value added in agricultural | % of GDP** | 6,03 | 13,6 | 11,2 | 13,47 | 10,6 | 11,99 | 13,6 | 20,11 |
| 15 | Mortality aged till 5 years | % | 2,99 | 3,92 | 4,36 | 3,08 | 3,79 | 3,93 | 3,89 | 9,95 |
| Cluster 4 | | | | | | | | | | |
| 16 | Quantity of childbirth by means of qualified medical personnel | % | 0,24 | 99,54 | 98,96 | 99,49 | 98,86 | 99,82 | 98,68 | 0,79 |
| 17 | Immunization against measles | % | 0,06 | 97,68 | 97,43 | 97,68 | 97,71 | 97,72 | 97,39 | 0,21 |
| 18 | The population per cent with primary education | % | 0,27 | 96,4 | 96,69 | 97,09 | 97,94 | 95,19 | 96,69 | 0,91 |
| 19 | Ratio of girls and boys in system of primary and secondary education | % | 0,34 | 98,28 | 97,71 | 98,79 | 98,02 | 99,29 | 97,29 | 1,14 |
| 20 | The improved water sources | % | 0,21 | 96,91 | 97,38 | 96,74 | 97,46 | 96,66 | 97,51 | 0,70 |
| 21 | The expected life expectancy at the birth | Years' quantity | 0,72 | 72,02 | 70,68 | 72,83 | 70,73 | 73,13 | 71,48 | 2,39 |
| 22 | The improved sanitation means | % | 1,21 | 74,96 | 71,38 | 74,86 | 70,08 | 74,65 | 74,45 | 4,03 |

* KOE – kilogram oil equivalent
** GDP – gross domestic product

Data from 1999 to 2011 were used for development of forecasting models. Data from 2011 to 2014 were used for forecasting of private TS on 3 steps forward.

Forecasting results of private TS on 3 steps forward, values of forecasting average relative errors for 3 steps forward, and also *AFER* (6) values (from 1999 to 2011) are given in Table 1.

The most essential influence on the development time of forecasting model on the base of SBT and MCSA is rendered by such MCSA parameters as iterations' number, size of antibodies' population, coefficient of antibodies' cloning and coefficient of clones' reproduction. In the reviewed example 600 iterations of MCSA for population of 20 antibodies were executed. Coefficient of antibodies' cloning was equal to 0,3. Coefficient of clones' reproduction was equal to 0,8.

Computer working under the 64-bit Windows 7 version with RAM of 2 Gb and the two-nuclear Pentium 4 processor with a clock frequency of 3,4 GHz was used for experiment. 115,5 seconds were spent for creation of one forecasting model. Thus, 462 seconds (7 minutes 42 seconds) are necessary for creation of 4 models, and 2541 seconds (42 minutes 21 seconds) are necessary for creation of 22 models, that in 5,5 times more.

At realization of the offered method the additional time-consuming caused by need of TS clustering procedure takes place. However, time spent for clustering procedure in the reviewed example takes only 0,254 seconds that much less time which needs to be spent for additional creation of 18 forecasting models [this time takes 2079 seconds (34 minutes 39 seconds)].

At application of this method for new grouped TS, that is for the first time, it is necessary to perform of TS clustering procedure for different quantity of clusters for finding of the optimum splitting determined by a minimum value of clustering algorithm criterion function (1). But also in this case essential decrease in time expenditure on forecasting for grouped TS takes place. Graphic dependences for elements' values of private TS, carried respectively to clusters 1, 2, 3 and 4, are presented in Fig. 2–5 (including forecast values for 3 steps forward).

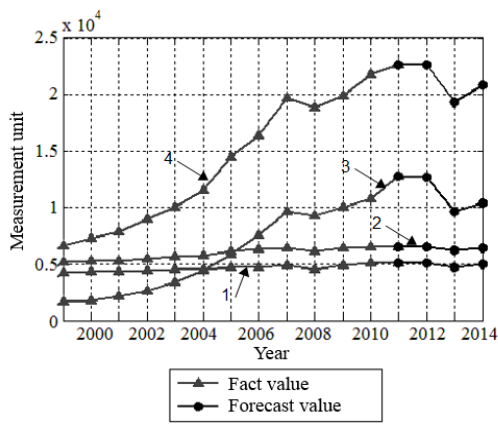

Fig. 2. Forecasting results for TS
of the cluster 1

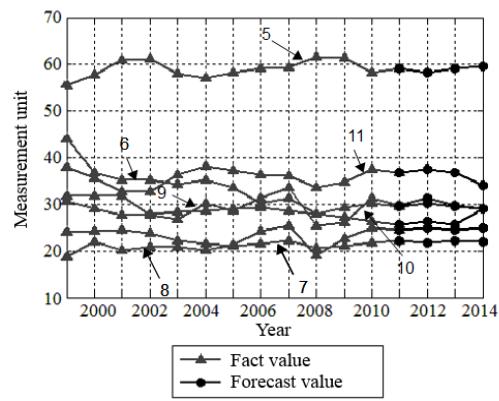

Fig. 3. Forecasting results for TS
of the cluster 2

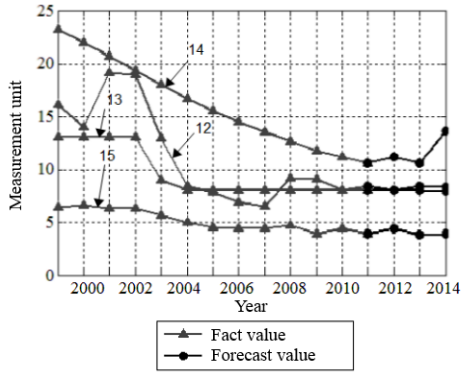

Fig. 4. Forecasting results for TS
of the cluster 3

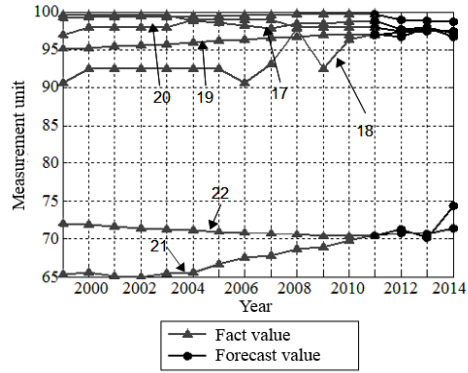

Fig. 5. Forecasting results for TS
of the cluster 4

All TS are numbered according to the numbering, given in Table 1. Designation «Measurement unit» on a vertical axis means measurement unit corresponding to private TS (the third column of Table 1).

For indicators with numbers 3, 4, 10, 11, 14 and 15 values of forecasting average relative error *AFER* (6) is significantly more, than for other indicators, and exceed 2%. In this regard the decision on improving of general forecasting models, corresponding to them, was made. As a result for indicators with numbers 3, 4, 10, 11, 14 and 15 individual forecasting models were received on the base of improved antibodies:

$$\_ + C \cdot C + \_ - S / E - E \cdot \_ \cdot Cb \_ bQfLeEaQgSaLaCi;$$
$$Q / E / Q - Q - E - E \cdot E - Q - SeLh \_ bSQdCeScEg \_ i;$$
$$L - E \cdot E \cdot \_ - C + C + E / S \cdot EhQc \_ hQbQgLeE?SeLc;$$
$$L \cdot Q \cdot C + 0 + C - E \cdot S - S + QhEiEhCeCdCcSbQgLe,$$
$$\_ + S \cdot Q + L + L + E \cdot S - C - ScEfEdCb \_ fQbEfLcLf;$$
$$Q + C / S \cdot Q \cdot C \cdot E / C \cdot C \cdot Ed \_ cSeCiEfEc \_ bQbCh,$$

in the form of respectively analytical dependences:

$$f_1(d^{j-1}, d^{j-2}, d^{j-3}, d^{j-4}, d^{j-5}, d^{j-6}) =$$
$$= \cos(\cos(\exp((\cos(d^{j-1}) \cdot \ln(d^{j-6})) \cdot \sin(d^{j-6})) + \exp(\mathrm{sqrt}(d^{j-2}) - \exp(d^{j-6})) \cdot$$
$$\cdot \sin(\ln(d^{j-4}) / \mathrm{sqrt}(d^{j-3}))) + d^{j-5} - \cos(d^{j-5}));$$
$$f_2(d^{j-1}, d^{j-2}, d^{j-3}, d^{j-4}, d^{j-5}, d^{j-6}, d^{j-7}) =$$
$$= \mathrm{sqrt}(\exp(\mathrm{sqrt}(\exp(\mathrm{sqrt}(d^{j-1} - \exp(d^{j-3})) - \sin(d^{j-6})) / \exp(\cos(d^{j-4}) \cdot$$
$$\cdot \mathrm{sqrt}(d^{j-5}))) / \exp(\sin(12,25) - d^{j-7})) -$$
$$- \mathrm{sqrt}(\ln(d^{j-2}) - \sin(d^{j-4})));$$

$f_3(d^{j-1}, d^{j-2}, d^{j-3}, d^{j-4}, d^{j-5}) =$

$= \ln(\exp(\exp(\exp(\sin(\ln(d^{j-4}) \cdot \sin(d^{j-3})) / \exp(129,43)) - \cos(\ln(d^{j-3}) +$

$+ \mathrm{sqrt}(d^{j-2}))) \cdot \cos(\mathrm{sqrt}(d^{j-5}) + d^{j-1})) \cdot \mathrm{sqrt}(d^{j-4}) - \exp(d^{j-1}))$ ;

$f_4(d^{j-1}, d^{j-2}, d^{j-3}, d^{j-4}, d^{j-5}, d^{j-6}, d^{j-7}) =$

$= \ln(\mathrm{sqrt}(\cos(\sin(\sin(\ln(d^{j-4}) + \mathrm{sqrt}(d^{j-3})) - \sin(d^{j-7})) \cdot \exp(\cos(d^{j-6}) \cdot$

$\cdot \cos(d^{j-5}))) \cdot \cos(\cos(d^{j-4}) - \exp(d^{j-2}))) + \exp(d^{j-1}) + \mathrm{sqrt}(d^{j-2}))$ ;

$f_5(d^{j-1}, d^{j-2}, d^{j-3}, d^{j-4}) =$

$= \sin(\mathrm{sqrt}(\sin(\cos(\ln(d^{j-1}) - \ln(d^{j-3})) - \exp(d^{j-1})) + \exp(\mathrm{sqrt}(d^{j-4}) \cdot d^{j-1}) \cdot$

$\cdot \ln(\cos(d^{j-4}) + \exp(d^{j-2}))) + \ln(\exp(d^{j-1}) + \sin(d^{j-3})))$ ;

$f_6(d^{j-1}, d^{j-2}, d^{j-3}, d^{j-4}, d^{j-5}, d^{j-6}, d^{j-7}) =$

$= \mathrm{sqrt}(\cos(\sin(\cos(\cos(\cos(d^{j-2}) \cdot \mathrm{sqrt}(d^{j-7})) \cdot d^{j-7}) +$

$+ \exp(\exp(d^{j-6}) / \exp(d^{j-3}))) / \cos(\cos(d^{j-1}) \cdot \sin(d^{j-4}))) \cdot \mathrm{sqrt}(d^{j-6} \cdot \exp(d^{j-5})))$ .

TS forecasting results on 3 steps forward, values of forecasting average relative errors for 3 steps forward, and also *AFER* (6) values (from 1999 to 2011) for indicators with numbers 3, 4, 10, 11, 14 and 15 are given in Table 2. Graphic dependences for elements' values of private TS for indicators with numbers 3, 4, 10, 11, 14 and 15 are presented in Fig. 6–11 (including forecast values for 3 steps forward with application of the general and improved general forecasting models).

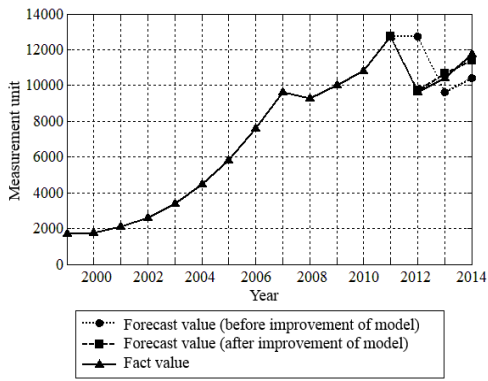

Fig. 6. Forecasting results for TS «Gross national income per capita on the Atlas method»

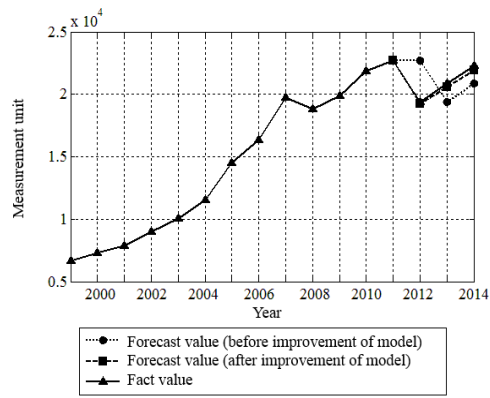

Fig. 7. Forecasting results TS «Gross national income per capita at par purchasing power»

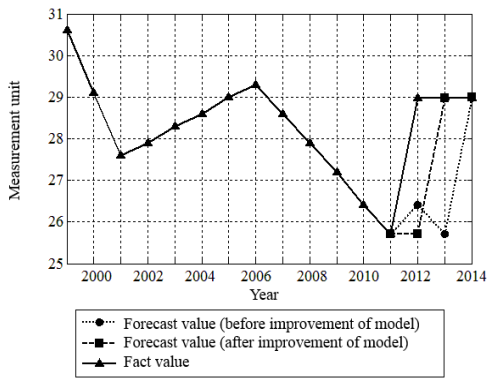

Fig. 8. Forecasting results for TS «Coefficient of teenage fertility»

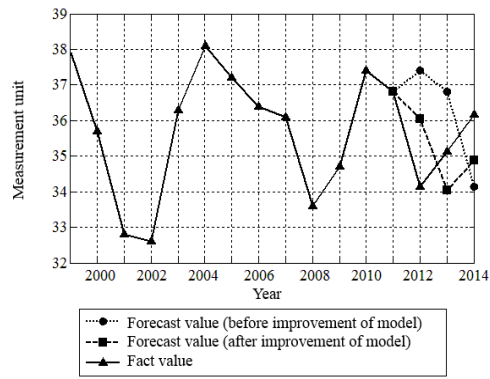

Fig. 9. Forecasting results for TS «Value added in the industry»

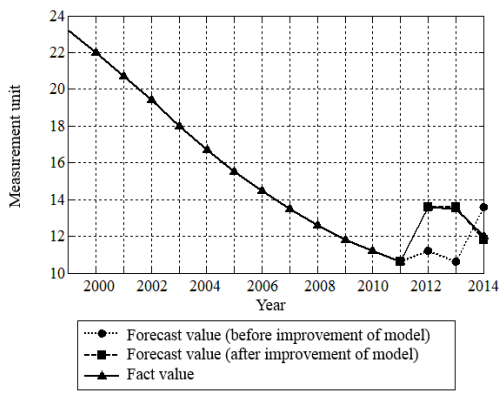

Fig. 10. Forecasting results for TS «Value added in agricultural»

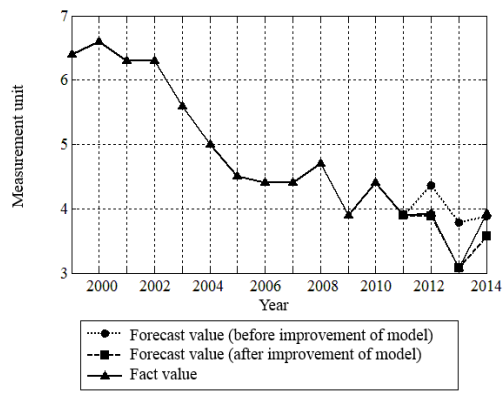

Fig. 11. Forecasting results for TS «Mortality aged till 5 years»

Apparently from Table 2 and Fig. 6–11 use of the improved general forecasting models allowed to improve forecasting results significantly. Thus calculation time increased for 693 seconds (11 minutes 33 seconds).

However in this case it was more than 2,2 times less, than in case of forecasting models' development for all 22 TS.

Table 2. The improved results of some macroindicators' forecasting

| No | Indicator name | Measurement unit | *AFER*, % | 2012 | | 2013 | | 2014 | | Error, % |
|----|----------------|------------------|-----------|------|------|------|------|------|------|----------|
| | | | | fact | forecast | fact | forecast | fact | forecast | |
| 3 | Gross national income per capita on the Atlas method | $ | 0,62 | 9642,28 | 9741 | 10406,16 | 10644 | 11740,99 | 11406 | 2,06 |
| 4 | Gross national income per capita at par purchasing power | $ | 0,49 | 19373,7 | 19238 | 20861,23 | 20568 | 22279,96 | 21895 | 1,29 |

Table 2. (Continued): The improved results of some macroindicators' forecasting

| 10 | Coefficient of teenage fertility | Births' quantity/ 1000 women | 1,43 | 28,97 | 25,7 | 29 | 28,97 | 28,98 | 29 | 4,30 |
|---|---|---|---|---|---|---|---|---|---|---|
| 11 | Value added in the industry | % of GDP** | 1,51 | 34,15 | 36,05 | 35,12 | 34,04 | 36,16 | 34,89 | 4,03 |
| 14 | Value added in agricultural | % of GDP** | 0,29 | 13,6 | 13,6 | 13,47 | 13,6 | 11,99 | 11,83 | 0,77 |
| 15 | Mortality aged till 5 years | % | 1,36 | 3,92 | 3,89 | 3,08 | 3,08 | 3,93 | 3,57 | 3,62 |

# 4 Conclusion

The offered approach of forecasting for grouped TS realizes the combined use of FCM-algorithm and forecasting models on the base of SBT and MCSA and provides individual forecast values for all TS in group with acceptable time-consuming.

The results of experimental studies received during macroeconomic indicators' forecasting of Russian Federation confirm prospects of application and further development for the offered approach.

Use of cluster analysis' algorithms allows to form clusters (subgroups) of the connected TS, having similar change laws of elements' values, and provides increase of TS forecasting speed. Application of the general forecasting models (forecasting models for TS-centroids of clusters) for private TS, entering into the relevant subgroups, doesn't lead to essential decrease in forecasting accuracy. Thus the demanded forecast accuracy for private TS can be reached in the course of general forecasting model improving with application of MCSA.